\documentclass[10pt,twocolumn]{extarticle}
\usepackage[top=1.9cm,bottom=1.9cm,left=1.5cm,right=1.5cm,columnsep=0.5cm]{geometry}
\usepackage{authblk}

\date{}
\usepackage[switch,mathlines]{lineno}
\usepackage{lmodern}
\usepackage{fix-cm}

\usepackage[numbers,sort&compress]{natbib}%
\makeatletter
\def\bibfont{\reset@font\fontfamily{\rmdefault}\normalsize\selectfont}%
\makeatother

\newlength\savewidth\newcommand\shline{\noalign{\global\savewidth\arrayrulewidth
\global\arrayrulewidth 1.5pt}\hline\noalign{\global\arrayrulewidth\savewidth}}
 
\usepackage{amsthm}%
\usepackage{abstract}

\usepackage{manyfoot}%
\usepackage{algorithm}%
\usepackage{algorithmicx}%
\usepackage{algpseudocode}%
\usepackage{listings}%
\usepackage[dvipsnames]{xcolor}
\usepackage{amsmath,amssymb,amsfonts}
\usepackage{graphicx}
\usepackage{textcomp}
\usepackage{booktabs,multirow}
\usepackage{hyperref}
\usepackage{pdfpages}
\usepackage{xr}
\usepackage[explicit]{titlesec}
\renewcommand{\thesubsection}{\thesection.\arabic{subsection}}
\renewcommand{\thesubsubsection}{\thesubsection.\arabic{subsubsection}}
\renewcommand{\theparagraph}{\thesubsubsection.\arabic{paragraph}}

\titleformat{\section}
  {\large\sffamily\bfseries}
  {\thesection.}
  {0.5em}
  {#1}
  []

\titleformat{name=\section,numberless}
  {\large\sffamily\bfseries}
  {}
  {0em}
  {#1}
  []

\titleformat{\subsection}[runin]
  {\sffamily\bfseries\fontsize{8}{10}\selectfont}
  {\thesubsection.}
  {0.5em}
  {#1. }
  []

\titleformat{\subsubsection}[runin]
  {\sffamily\small\bfseries\itshape}
  {\thesubsubsection.}
  {0.5em}
  {#1. }
  []

\titleformat{\paragraph}[runin]
  {\sffamily\small\itshape\bfseries}
  {\indent\theparagraph. }
  {0em}
  {#1.}

\titlespacing*{\section}{0pc}{3ex plus 4pt minus 3pt}{5pt}
\titlespacing*{\subsection}{0pc}{2.5ex plus 3pt minus 2pt}{6pt}
\titlespacing*{\subsubsection}{0pc}{2ex plus 2.5pt minus 1.5pt}{6pt}
\titlespacing*{\paragraph}{0pc}{1.5ex plus 2pt minus 1pt}{6pt}

\usepackage[labelfont={bf,sf},
            labelsep=period,
            figurename=Fig.]{caption}
\usepackage{booktabs}
\usepackage{etoolbox}

\DeclareCaptionFormat{pnasformat}{\normalfont\sffamily\fontsize{7}{9}\selectfont#1#2#3}
\AtBeginEnvironment{tabular}{\sffamily\fontsize{7.5}{10}\selectfont}
\AtBeginEnvironment{tabular*}{\sffamily\fontsize{7.5}{10}\selectfont}

\makeatletter
\renewcommand\tagform@[1]{\maketag@@@{\bfseries[\ignorespaces #1\unskip\@@italiccorr]}}
\makeatother

\providecommand{\refdoi}[1]{\urlstyle{rm}\url{#1}}              

\begin{document}

\title{Prediction of Rectal Cancer Regrowth from Longitudinal Endoscopy}

\author[a, b, 1]{Jorge Tapias Gomez}
\author[c]{Despoina Kanata}
\author[a]{Aneesh Rangnekar}
\author[c]{Christina Lee}
\author[c]{Hannah Williams}
\author[c]{Hannah Thompson}
\author[c]{J. Joshua Smith}
\author[d]{Francisco Sanchez-Vega}
\author[e, f]{Mert R. Sabuncu}
\author[c]{Julio Garcia-Aguilar*}
\author[a]{Harini Veeraraghavan*}

\affil[a]{Department of Medical Physics, Memorial Sloan Kettering Cancer Center, New York, NY}
\affil[b]{School of Computer Science, Cornell University and Cornell Tech, New York, NY}
\affil[c]{Department of Surgery, Colorectal Service, Memorial Sloan Kettering Cancer Center, New York, NY}
\affil[d]{Department of Epidemiology and Biostatistics, Memorial Sloan Kettering Cancer Center, New York, NY}
\affil[e]{Department of Radiology, Weill Cornell Medical College, New York, NY}
\affil[f]{School of Electrical and Computer Engineering, Cornell University and Cornell Tech, New York, NY}

\affil[1]{To whom correspondence may be addressed. Email: tapiasj@mskcc.org}
\affil[*]{Equal senior author contribution.}

\twocolumn[
  \maketitle
  \begin{onecolabstract}
  Clinical trials have shown the benefit of watch-and-wait (WW) surveillance for patients with rectal cancer showing a complete or near-complete clinical response (CR) directly after treatment (restaging). However, there are no objective and accurate methods to detect local tumor regrowth (LR) early in patients undergoing WW from follow-up exams. Hence, we developed Temporal Rectal Endoscopy Cross-attention (TREX), a longitudinal deep learning approach that combines pairs of images acquired at restaging and follow-up to distinguish CR from LR. TREX uses pretrained Swin Transformers in a siamese setting to extract features from longitudinal images and dual cross-attention to combine the features without spatial co-registration between image pairs. TREX and Swin-based baselines were trained under two settings: (a) detecting LR or CR at the last available follow-up and (b) early detection of LR at 3--6, 6--12, and 12--24 months before clinical confirmation. TREX achieved the highest accuracy in detecting LR with a high sensitivity of 97\% $\pm$ 6\% and a balanced accuracy of 90\% $\pm$ 3\%, and outperformed all baselines in early detection at both 3--6 (74\% $\pm$ 1\%) and 6--12 months (62\% $\pm$ 4\%) prior to clinical detection. Clinical validation via a surgeon survey showed that TREX matched clinician-level overall accuracy (TREX: 86.21\% versus clinicians: 87.84\% $\pm$ 1.28\%). Finally, we explored TREX's ability to predict treatment response by combining pre-treatment (pre-TNT) and restaging endoscopies, achieving a balanced accuracy of 73\% $\pm$ 12\%. These results show that longitudinal deep learning analysis of endoscopy may improve surveillance and enable earlier identification of rectal cancer regrowth.
  \end{onecolabstract}
  \vspace{1em}
]

\section{Introduction}

\begin{figure*}[!t]
    \includegraphics[width=\linewidth]{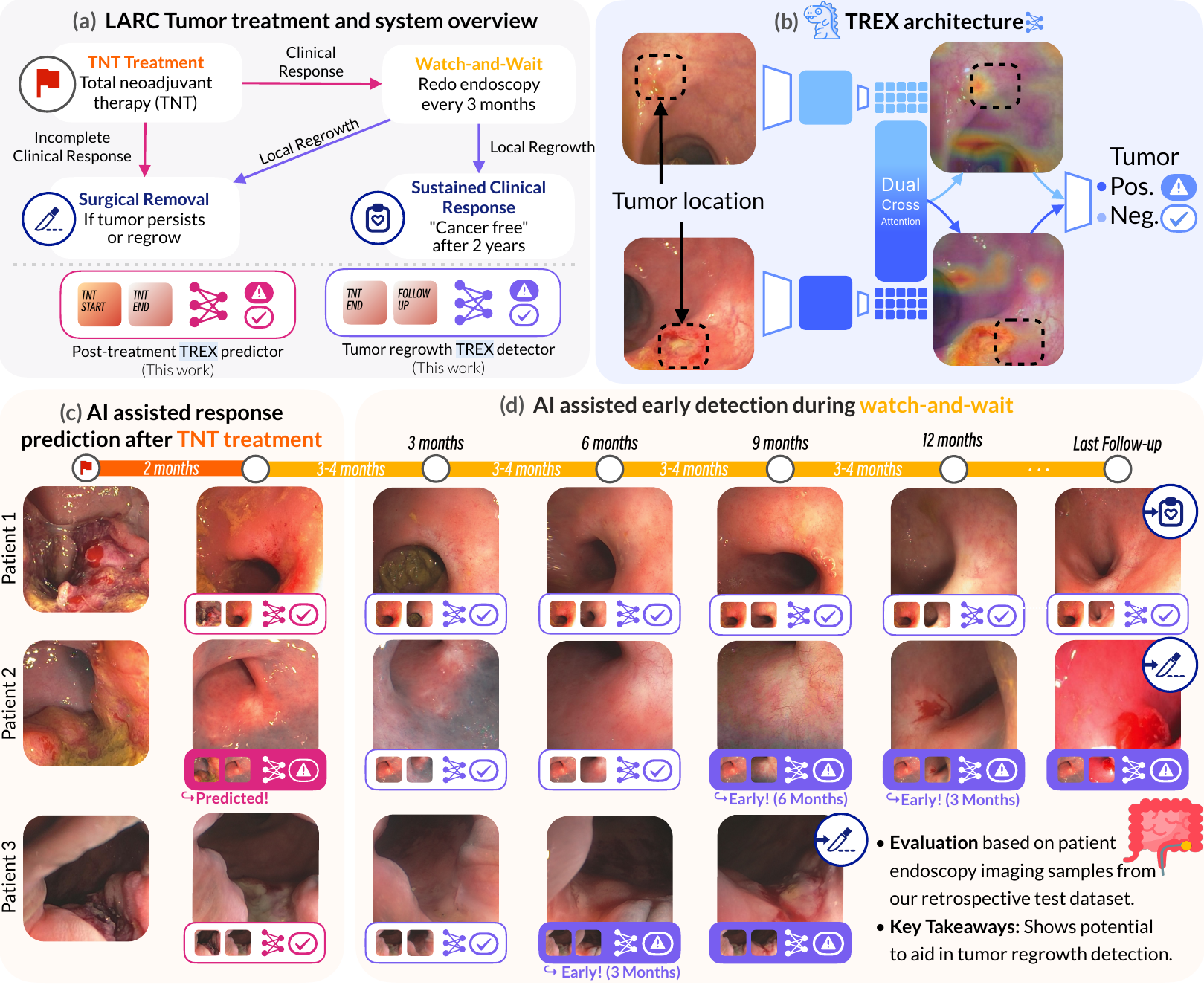}
    \caption{Prediction and surveillance system (TREX) for locally advanced rectal cancer under total neoadjuvant therapy (TNT) and watch-and-wait (WW) management. (a) Clinical workflow: patients undergo TNT treatment followed by either surgical resection for persistent/recurrent disease or WW surveillance with endoscopy every 3 months, achieving a complete or near-complete clinical response (CR) at 2 years. (b) TREX architecture employs a dual cross-attention mechanism applied to paired endoscopic images producing a binary tumor-positive/negative classification. TREX is fine-tuned on two tasks: post-treatment response prediction from pre-treatment (pre-TNT) and restaging images (see part c) and follow-up regrowth detection from WW follow-ups (see part d). (c, d) Representative endoscopic image sequences from three patients: Patient 1 achieves CR across all timepoints; Patient 2 demonstrates early regrowth flagged up to 6 months ahead of clinical confirmation; Patient 3 shows regrowth detection 3 months early. 
    }
    \label{fig:problemOverview}
\end{figure*}

There is growing evidence that some rectal adenocarcinomas can be cured without surgery~\cite{langenfeld2024american}. Given the rising number of younger people diagnosed with rectal cancer, treatment strategies that preserve patients' quality of life while not compromising survival outcomes are increasingly needed. The Organ Preservation in Rectal Adenocarcinoma (OPRA) clinical trial demonstrated that 47\% of patients treated with total neoadjuvant therapy (TNT), consisting of neoadjuvant chemotherapy and chemoradiation, preserved their rectum through a watch-and-wait (WW) strategy (Fig.~\ref{fig:problemOverview}a). Importantly, this organ-preservation approach did not compromise patients’ chance of cure~\cite{OPRAH}. A detailed overview of the treatment regimen is provided in \textit{SI Appendix}~\ref{SI-Fig:LARC_Overview}.


However, despite showing a complete or near-complete clinical response (referred to as CR) at the restaging exam performed four to twelve weeks after total neoadjuvant therapy, 33\% of the same patients developed local regrowth (LR) at the tumor site and proceeded to surgery~\cite{LongTermOprah, dattani2018oncological, CHADI2018825}. Early detection of local regrowth is thus essential for successful salvage surgery and preventing metastatic spread. Patient selection for watch-and-wait strategy is based on digital rectal examination, flexible sigmoidoscopy or endoscopic imaging, and magnetic resonance imaging. As most local regrowths involve the rectal wall and can be seen from the inside of the rectum, endoscopy is considered the primary tool to select patients for watch-and-wait strategy~\cite{williams2024endoscopic}. However, visual endoscopic assessment is challenged by large inter-observer variation (as high as 37\%), due in part to differences in clinician experience with watch-and-wait strategy~\cite{felder2021}. Subjective assessment is also only 68\% accurate  for detecting residual tumor after TNT~\cite{felder2021}. These limitations highlight the need for objective and accurate methods to distinguish sustained CR from LR during surveillance. Particularly, as interest in watch-and-wait expands outside of highly specialized centers~\cite{maas2015assessment, Ko2019, Paardt_SystematicReview, Kawai2017, Safont2024}.

In recent years, deep learning (DL) models have shown remarkable capabilities across medical imaging tasks for radiographic modalities such as computed tomography (CT) and magnetic resonance imaging (MRI). Applications range from image registration~\cite{WANG2023102962,JuePRoRSeg} and cancer detection~\cite{BreastCancer,LungCancerScreening} to longitudinal analysis for predicting and detecting treatment responses across various cancerous~\cite{Chen2025IJROBP_PretreatmentMidtreatmentCT,Ke2023DisColonRectum_DeepRPRC,Jin2021NatCommun_LongitudinalRPNet,Sun_LOMIAT_MICCAI2024} and non-cancerous conditions~\cite{Gerig2016MedImageAnal_LongitudinalModeling,LILAC_heejong}.

\begin{figure*}[t]
    \centering
    \includegraphics[width=0.98\linewidth]{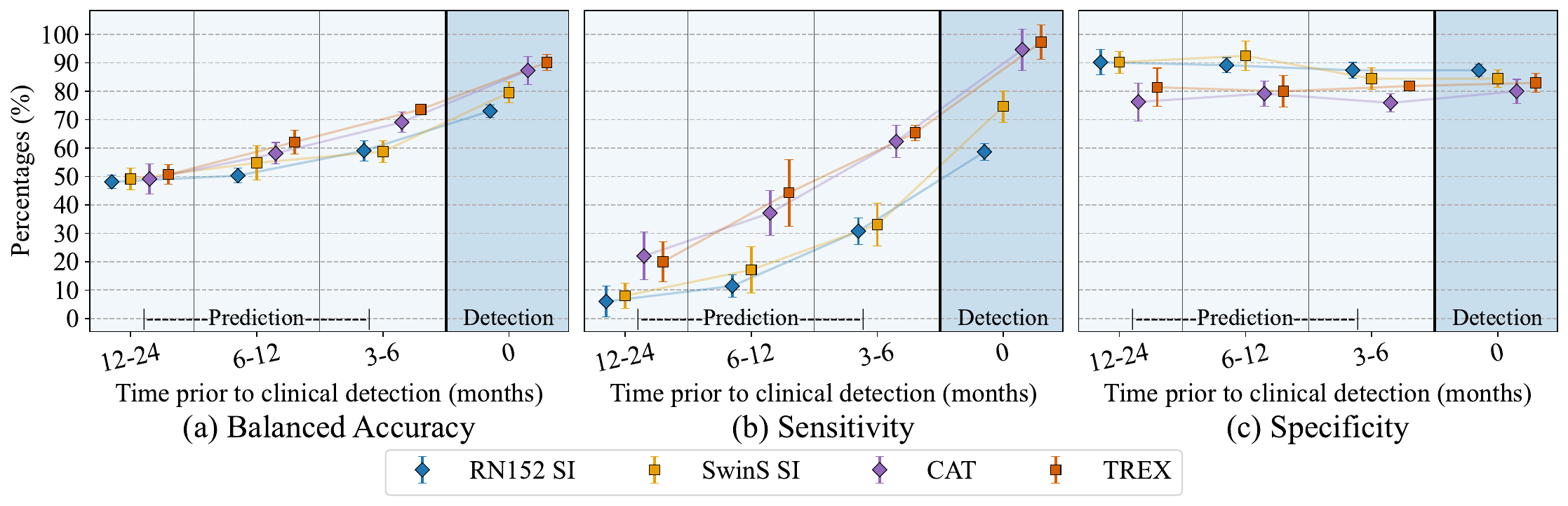}
    \caption{Performance of the baseline models and TREX models across longitudinal follow-up timepoints, where clinicians labeled the image at the last available follow-up (timepoint 0) and any other previous follow-ups are retrospectively assigned that label. TREX achieved the highest balanced accuracy and sensitivity at clinically relevant timepoints, particularly 3–6 months before clinical detection and at the final follow-up (0 months), outperforming single-image models (RN152-SI, SwinS-SI) and the conventional siamese baseline (CAT). Earlier timepoints (12–24 months before detection) remained challenging for all models, with performance near chance levels. Balanced accuracy, sensitivity, and specificity were calculated for each fold by averaging across all subjects with available follow-up images. Timepoint labels are abbreviated for readability: `0' = clinical detection at the last follow-up, `3--6' = 3--6 months before detection, `6--12' = 6--12 months before detection, and `12--24' = 12--24 months before detection.}
    \label{longitudinal_ID}
\end{figure*}

In contrast, DL models for endoscopic image analysis remain relatively nascent for post-treatment scenarios, particularly the watch-and-wait strategy. Most existing work has focused on detecting and segmenting pre-cancerous lesions such as polyps~\cite{PolypDetect,yamada2019,ALI2021102002,fan2020pranet,Dong2023PolypPVT, Byrne94, JIN20202169}, with additional studies addressing non-cancerous conditions~\cite{turan2022}, tumor detection prior to treatment~\cite{Dong2023, Almeida2025}, and synthetic image generation~\cite{Lin2026StyleGANPolypGeneralization,Golhar2025LesionClassificationGAN}. More recent efforts have explored foundation models for endoscopic applications~\cite{Wang2023EndoFM,Paderno2024Laryngoscope_CVFoundationModels,Jong2025GastroNet5M}. In comparison, relatively little work has focused on developing models for post-treatment tumor assessment, despite its greater clinical complexity~\cite{thompson2023,Haak_Endo_ClinicVariables}. Models developed for this task have shown promising performance~\cite{Williams2024}, although their accuracy remains limited for detecting tumor regrowth, which is more difficult due to differences in appearance and shape compared to pre-treatment tumors. To the best of our knowledge, our study is the first to leverage longitudinal endoscopic image pairs to predict and detect rectal cancer regrowth during watch-and-wait surveillance.

One likely reason for this gap is that most longitudinal DL models assume some degree of spatial alignment between images or rely on regions of interest enclosing the relevant anatomy~\cite{Jin2021NatCommun_LongitudinalRPNet,Sun_LOMIAT_MICCAI2024}. Such assumptions are impractical in endoscopy where images are acquired across different timepoints spanning several months, with large perspective changes arising naturally from variability in scope positioning and transient visual artifacts such as the presence of stool, blood, telangiectasia, and specular reflections~\cite{thompson2023}. Recent works~\cite{DCAT, CrossAttentionChangeDetection} have demonstrated that spatially unaligned image pairs can still be effectively leveraged for longitudinal analysis. In particular, several methods have proposed the application of cross-attention mechanisms where features from one timepoint explicitly attend to corresponding features from the other, enabling temporal interaction during feature extraction rather than only at late fusion. This mechanism has also been extended to multimodal longitudinal domains such as pairing chest X-rays with prior radiology reports to model temporal disease progression~\cite{zhu2023utilizing}. Overall, these findings make cross-attention well suited to longitudinal endoscopy, where images acquired at different timepoints are rarely spatially aligned but may share diagnostically relevant patterns.

Building on these insights, we hypothesize that cross-attention between longitudinal images can improve local regrowth detection by capturing feature correspondences across timepoints. However, cross-attention alone does not account for the temporal interval between image acquisitions, limiting its ability to interpret important changes occurring over different timepoints. Furthermore, prior clinical studies have shown that temporal information for WW management is important as the chances for disease-free survival improve the longer the patient is in the watch-and-wait period. Hence, we explicitly encode the temporal interval between acquisitions to add clinically relevant context. The elapsed time between examinations can provide important clinical context that visual features alone may not capture.

Integrating these components, we propose TREX (Temporal Rectal Endoscopy Cross-Attention), a longitudinal deep learning framework (Fig.~\ref{fig:problemOverview}b) that combines cross-attention–based feature integration with explicit temporal encoding to analyze paired endoscopic images for detection and early prediction of local regrowth.

As shown in Fig.~\ref{fig:problemOverview}c and Fig.~\ref{fig:problemOverview}d, after the initial assessment at restaging (8 $\pm$ 4 weeks), patients with a complete or near-complete CR who undergo WW management have an endoscopic follow-up every 3--4 months for the first 2 years and every 6 months for an additional 3 years. This results in multiple longitudinal images per patient, albeit with a larger number of examples for CR than LR. We therefore evaluated whether our approach can identify LR both at clinical detection and at earlier timepoints, including at restaging. Specifically, we measured performance at multiple intervals (3--6, 6--12, and 12--24 months) prior to clinical detection, as well as at the final follow-up.

Our contributions are: (1) a siamese transformer architecture with a bidirectional cross-attention to extract informative features from image pairs without requiring explicit spatial alignment, (2) explicit temporal information encoding to enhance the model's predictive ability, and (3) evaluation of the clinical relevance of the proposed models by comparing their performance with that of clinicians at the final timepoint.

\begin{table}[!t]
    \centering
    \def\arraystretch{1.25}
    \caption{Response prediction performance. All metrics are reported as mean $\pm$ standard deviation (\%) across cross-validation folds. SwinS-SI pre-TNT predicts using only pre-TNT images, SwinS-SI uses only restaging images, and TREX uses paired pre-TNT and restaging images.}
    \resizebox{0.98\linewidth}{!}{%
    \begin{tabular}{llll}
    Models & Bal. accuracy & Sensitivity & Specificity \\
    \shline
    SwinS-SI pre-TNT & 58.90 $\pm$ 10.89 & 72.29 $\pm$ 17.87 & 43.23 $\pm$ 14.11 \\
    SwinS-SI & 61.41 $\pm$ 8.33 & 58.10 $\pm$ 11.56 & 64.85 $\pm$ 19.69 \\
    TREX & 73.22 $\pm$ 8.83 & 84.60 $\pm$ 8.85 & 61.23 $\pm$ 13.30 \\
    \bottomrule
    \end{tabular}}
    \label{tab:PredictingWW}
\end{table}

\section{Results}

\subsection{Models combining temporal images improve performance over single-image models}

\subsubsection{Detecting and predicting local regrowth at follow-up}

The performance of the models across early prediction timepoints and at the clinical detection timepoint is shown in Fig.~\ref{longitudinal_ID}. TREX achieved the highest balanced accuracy and sensitivity at the 6--12 months, 3--6 months, and 0-months timepoints. At the later evaluation timepoints closest to clinical detection (3--6 months and 0 months, n = 104), TREX achieved significantly higher accuracy than SwinS-SI (odds ratio: 3.80, p$=$0.001), RN152-SI (n = 104, odds ratio: 4.60, p$=$0.008), and CAT (odds ratio: 5.00, p$=$0.043).

\begin{itemize}
    \item \textbf{12--24 months before detection:} At this very early timepoint, all models performed near chance level, with balanced accuracy around 50\%, largely because sensitivity remained below 20\% for all models.
    \item \textbf{6--12 months before detection:} Balanced accuracy of all models improved relative to the early 12--24 months timepoint, with TREX's performance increasing by 12\% and reaching 62\% $\pm$ 4\%. TREX and CAT achieved higher sensitivity (TREX: 44\% $\pm$ 12\%, CAT: 37\% $\pm$ 8\%) compared to the single-image models (SwinS-SI: 17\% $\pm$ 8\%, RN152-SI: 11\% $\pm$ 4\%) (Fig.~\ref{longitudinal_ID}b) but lower specificity (Fig.~\ref{longitudinal_ID}c). These findings suggest that early follow-up images remain difficult to distinguish visually between patients with CR and those who later develop LR.
    \item \textbf{3--6 months before detection:} Accuracy of all models improved, with TREX's balanced accuracy increasing by 12 percentage points, reaching 74\% $\pm$ 1\%, relative to the 6--12 months timepoint. TREX and CAT again achieved higher sensitivity than the single-image models, with only a modest reduction in specificity. The gap in sensitivity between TREX and single-image models increased at this timepoint, whereas the gap in specificity narrowed (Fig.~\ref{longitudinal_ID}b and Fig.~\ref{longitudinal_ID}c). 
    \item \textbf{Clinical detection timepoint (0 months):} The balanced accuracy of all models improved at clinical detection. TREX achieved the highest balanced accuracy at this timepoint (90\% $\pm$ 3\%). The sensitivity advantage of TREX over the single-image models increased further, while differences in specificity became smaller.
\end{itemize}

\begin{figure}[!t]
    \centering
    \includegraphics[width=0.98\linewidth]{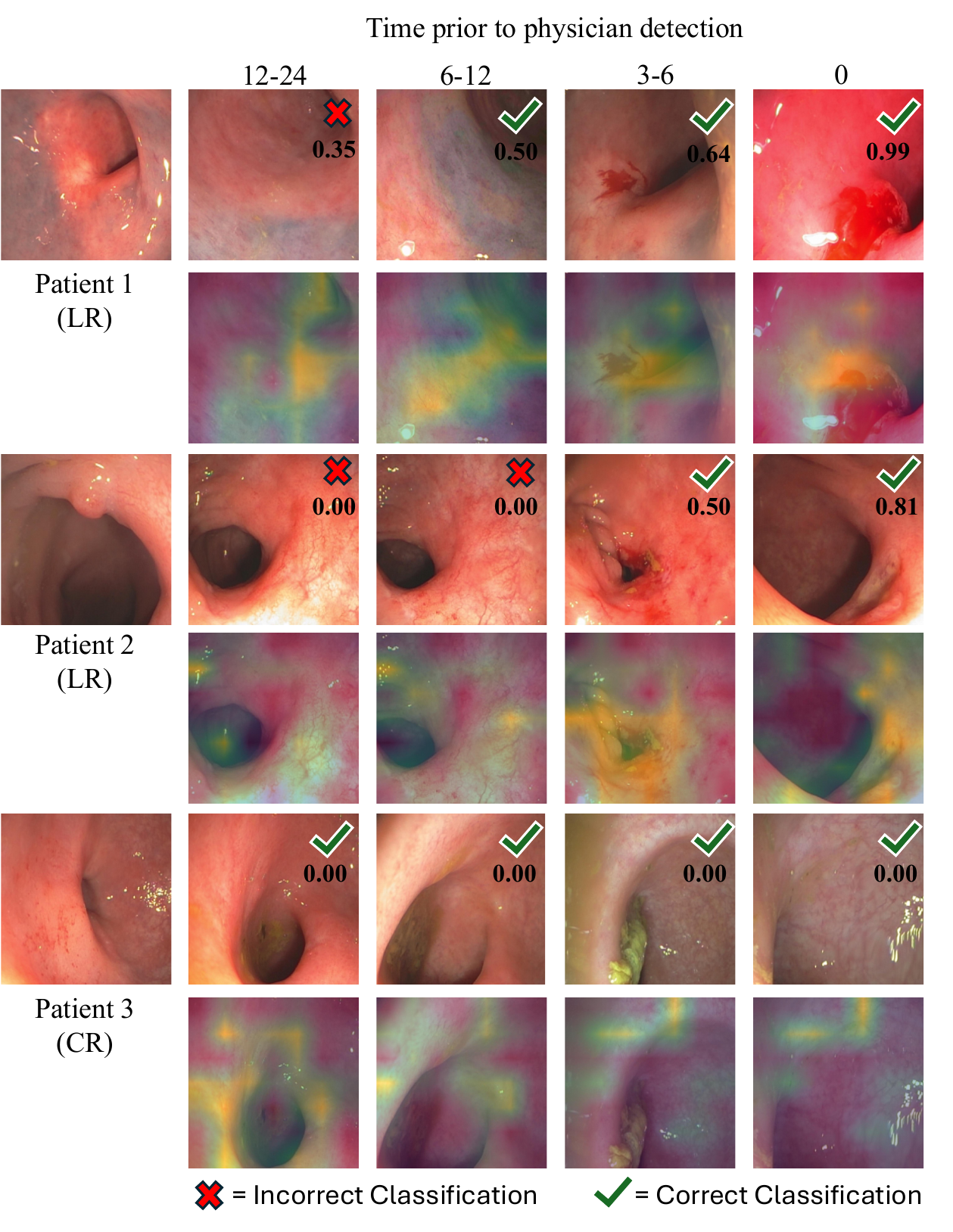}
    \caption{Representative trajectories of three patients during the WW period. Patients 1 and 2 developed local regrowth, with TREX correctly identifying tumor presence as early as 3--6 months prior to clinical detection in both cases, and 6--12 months prior in Patient 1. Patient 3 achieved a sustained CR, which TREX correctly classified throughout the entire WW period.}
    \label{fig:PatientTimeline}
\end{figure}

Representative patient trajectories illustrating TREX's predictions during the watch-and-wait period are shown in Figure~\ref{fig:PatientTimeline}. A detailed per-timepoint breakdown for the siamese models is provided in the \textit{SI Appendix}, Table~\ref{SI-tab:all_times_model_performance} and for the single-image models in \textit{SI Appendix}, Table~\ref{SI-tab:model_performance}. We also provide UMAP visualization of the feature separation in \textit{SI Appendix}, Fig.~\ref{SI-Fig:UMAP} and PCA inter-cluster distances \textit{SI Appendix}, Table~\ref{SI-tab:SI_UMAP_InterClusterDistances}, where TREX had the highest inter-cluster distance at 12--24, 3--6 and 0 months before clinical detection.

\subsubsection{Predicting post-treatment response at restaging}

\begin{figure}[!t]
    \centering
    \includegraphics[width=\columnwidth]{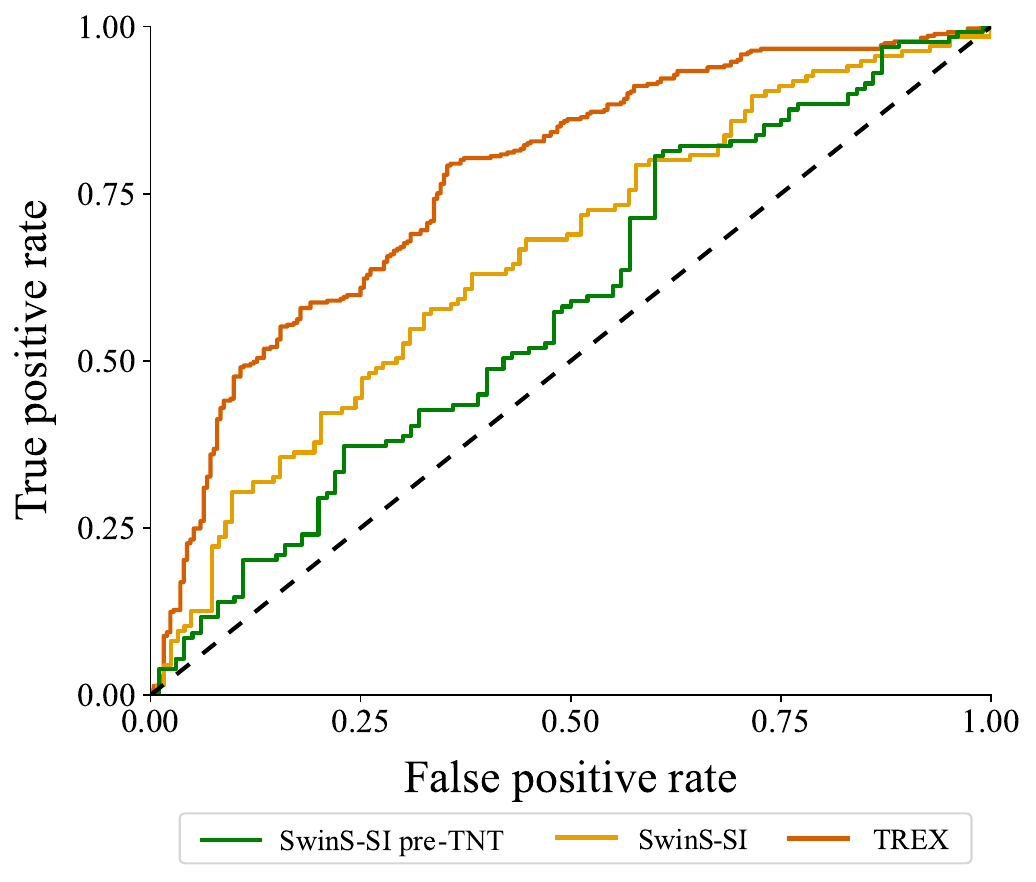}
    \caption{ROC curves for prediction of LR versus CR at the restaging timepoint. TREX, which combines paired pre-treatment (pre-TNT) and restaging endoscopic images, achieved higher balanced accuracy, sensitivity, and AUC than the pre-TNT-only model (SwinS-SI pre-TNT), demonstrating the value of longitudinal image pairing. Performance was comparable to the restaging-only model (SwinS-SI).}
    \label{PredictingWW_AUC}
\end{figure}

The goal of this experiment was to assess the benefit of combining image pairs rather than using single images to predict LR at the earliest timepoint, namely at restaging. TREX was trained with paired pre-TNT and restaging images. For comparison, two different instances of SwinS-SI were created using only pre-TNT endoscopic images (SwinS-SI pre-TNT) and using both pre-TNT and restaging endoscopic images (referred to as SwinS-SI) to predict LR versus CR. Table~\ref{tab:PredictingWW} and Fig.~\ref{PredictingWW_AUC} show that TREX achieved higher balanced accuracy (73\%), sensitivity (84\%), and AUC (0.77) than SwinS-SI pre-TNT (odds ratio: 2.75, $p=0.0176$). However, its performance was statistically similar to SwinS-SI (odds ratio: 2.2, $p=0.05$), indicating only marginal benefit from combining the pre-TNT and restaging images for predicting subsequent regrowth in patients who already showed visible clinical response and were thus placed on WW surveillance. 

\begin{table}[!t]
    \centering
    \def\arraystretch{1.25}
    \caption{Clinical validation on the endoscopic image survey (n = 58), reported as percentages. TREX achieved overall accuracy comparable to clinicians while maintaining perfect specificity (100\%) for clinical response (CR) cases, reducing the risk of unnecessary surgery. Clinician values represent mean $\pm$ standard deviation across survey participants. $^{\dagger}$CAT and TREX results include evaluation across all possible prior image-pair combinations corresponding to the survey cases (n = 772).}
    \resizebox{0.98\linewidth}{!}{%
    \begin{tabular}{llll}
    Evaluator & Accuracy & Sensitivity & Specificity\\
    \shline
        Clinicians          & 87.84 $\pm$ 1.28 & 79.32 $\pm$ 3.73 & 95.25 $\pm$ 2.64 \\
    \midrule        
        SwinS-SI            & 75.86 & 62.96 & 87.10 \\
        CAT                 & 82.76 & 62.96 & 100.00 \\
        TREX                & 86.21 & 70.37 & 100.00 \\
    \midrule 
        CAT$^{\dagger}$     & 81.03 & 74.07 & 87.10 \\
        TREX$^{\dagger}$    & 86.21 & 77.78 & 93.55 \\
    \bottomrule
    \end{tabular}}
    \label{tab:ClinicalSurvey}
\end{table}

\subsection{Clinical validation}

The average performance across all clinician participants, including general surgery residents (n = 4), colorectal surgery fellows (n = 7), and colorectal surgeon attendings (n = 5), is shown in Table~\ref{tab:ClinicalSurvey}, with a breakdown by clinician group provided in the \textit{SI Appendix} Table~\ref{SI-tab:SIClinicalSurvey}. On images containing LR, TREX correctly identified 70\% compared to 79\% for clinicians. On images with CR, the same TREX model correctly identified 100\% compared to 95\% for clinicians. TREX was more accurate than either CAT or SwinS-SI models on these survey images. 

We additionally evaluated TREX and CAT using all combinations of image pairs for the survey images. Under this expanded evaluation, TREX showed sensitivity of 78\%, closely matching the 79\% achieved by clinicians and exceeding the 74\% of CAT. Although specificity of both models decreased relative to clinicians, TREX maintained specificity above 90\%.

\begin{figure}[!t]
\centerline{\includegraphics[width=\columnwidth]{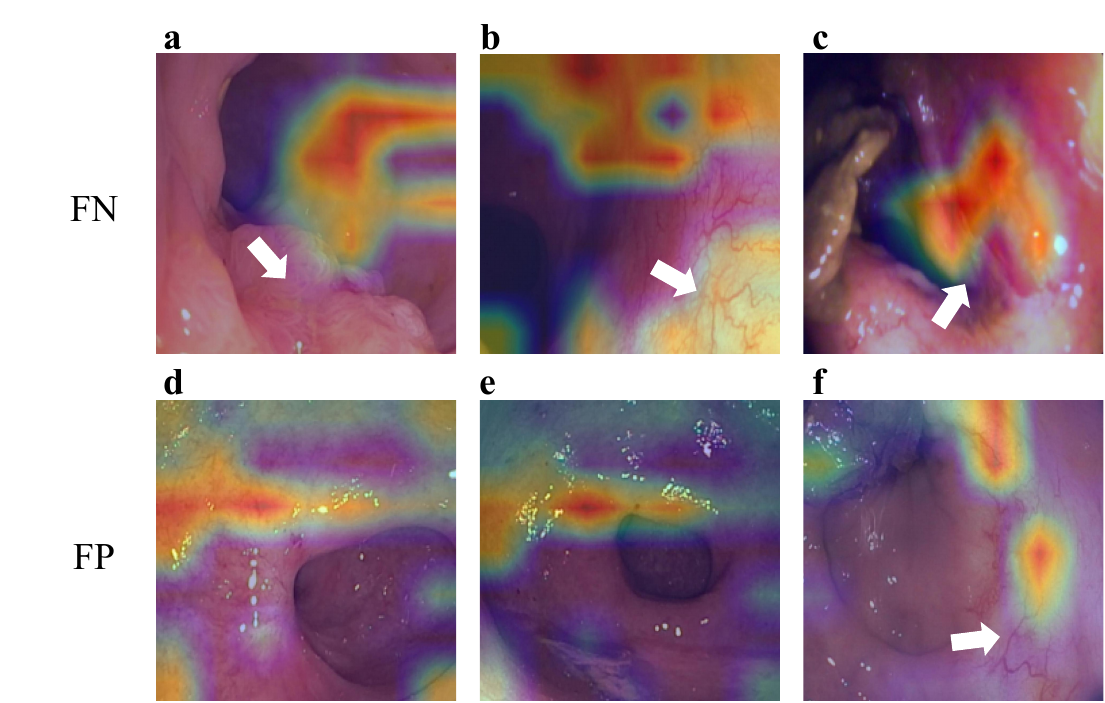}}
\caption{Representative Grad-CAMs for misclassified survey images. The top row shows false negatives, with white arrows indicating regions of residual disease. From left to right: (a) nodules, (b) vascular abnormalities, and (c) nodules with stool and partial visualization of the scope. The bottom row shows false positives: (d) normal mucosal fold, (e) stool, and (f) telangiectasia.}
\label{fig:gradCAMs}
\end{figure}

We further generated the Gradient-weighted Class Activation Mapping (Grad-CAM) visualizations~\cite{Selvaraju2020} for the misclassified cases and reviewed them with a clinician to diagnose TREX failure modes. Fig.~\ref{fig:gradCAMs} shows representative examples wherein TREX incorrectly detected LR in 15 CR images despite no visible tumors. The false positives resulted from misidentifying telangiectasia (n = 6; 40.0\%), normal mucosal folds (n = 4; 26.6\%), and artifacts or poor image quality (n = 4; 26.6\%). TREX also failed to identify LR in 13 images with visible tumor. Among these false negatives, the most common confounders were nodules (n = 5; 38.4\%), followed by ulcer (n = 3; 23.1\%), vascular abnormalities (n = 2; 15.4\%), blood (n = 2; 15.4\%), visualization of the scope (n = 2; 15.4\%) and stool (n = 1; 7.6\%). All clinical survey false negatives and false positives with their overlaid Grad-CAMs can be found in the \textit{SI Appendix}, Fig.~\ref{SI-Fig:FN_Survey} and Fig.~\ref{SI-Fig:FP_Survey}.

\begin{figure*}[t]
    \centering
    \includegraphics[width=0.98\linewidth]{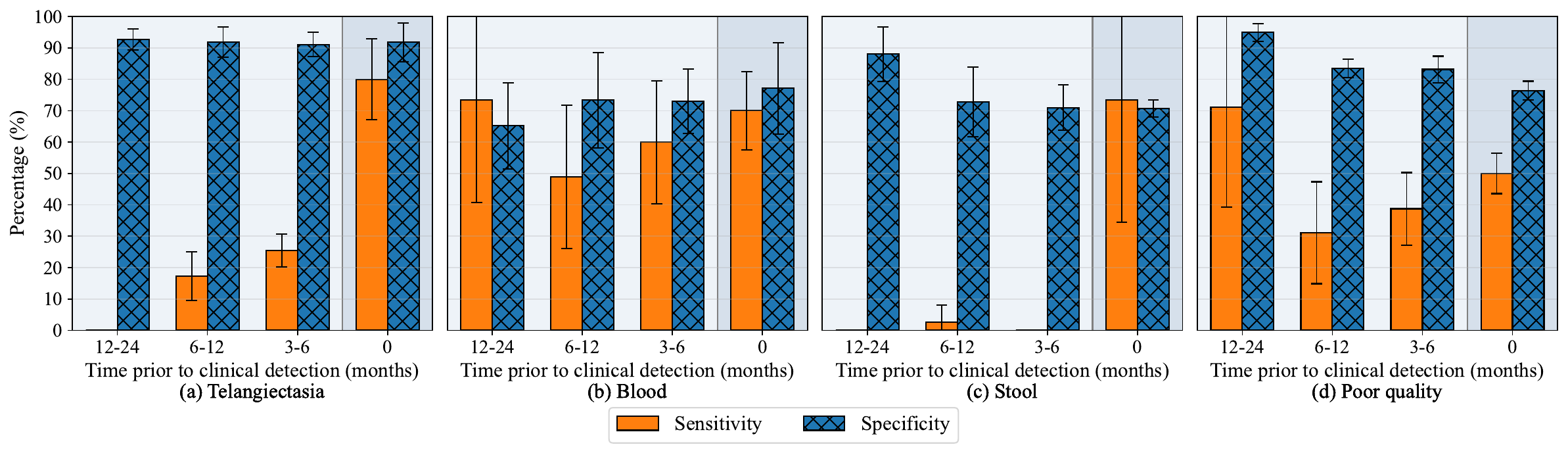}
    \caption{TREX specificity and sensitivity on common endoscopic image artifacts for the analyzed timepoints (averaged over folds). Timepoint labels are abbreviated for readability: `0' = clinical detection at the last follow-up, `3--6' = 3--6 months before detection, `6--12' = 6--12 months before detection, and `12--24' = 12--24 months before detection.}
    \label{fig:dataset_confounders}
\end{figure*}

\begin{figure}[!t]
    \centering
    \includegraphics[width=0.98\linewidth]{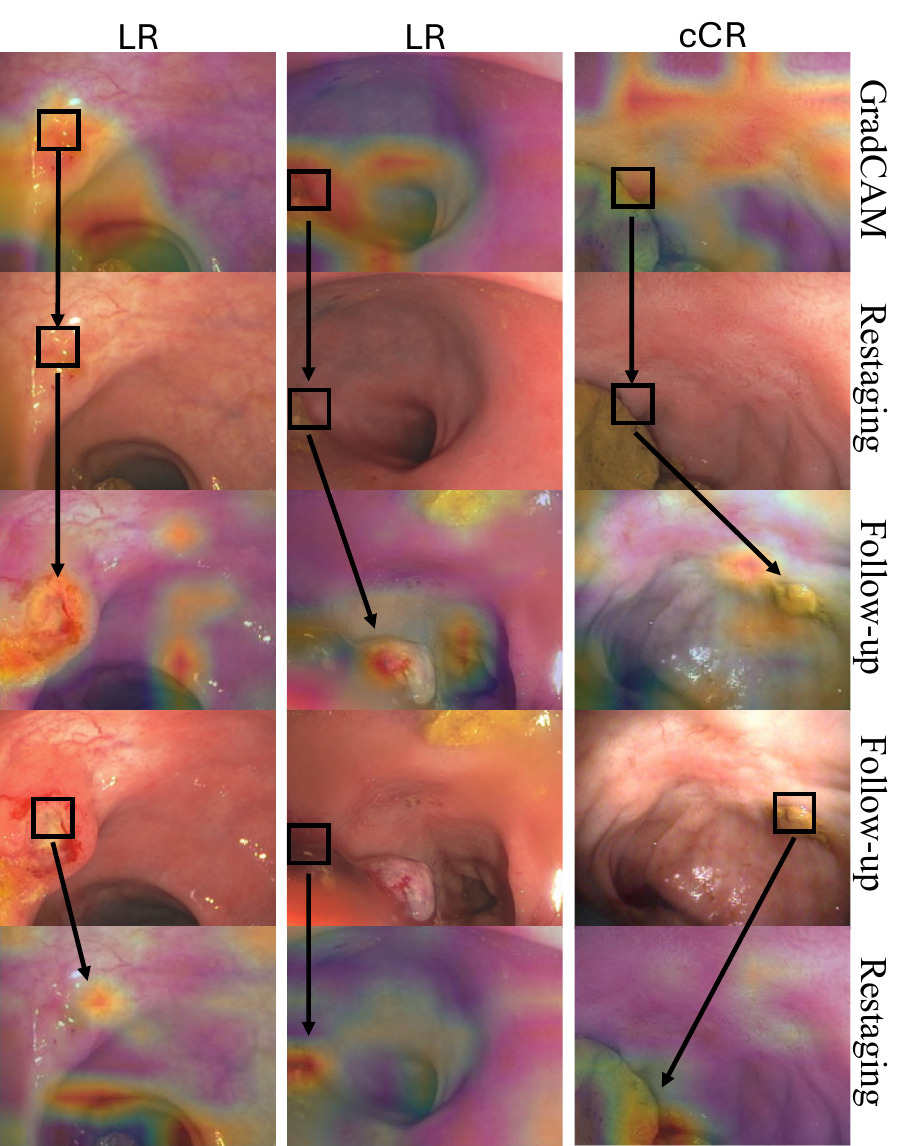}
    \caption{Grad-CAM and attention maps for four representative test cases produced by TREX, illustrating good correspondence of relevant spatial features between image pairs.}
    \label{Attention}
\end{figure}

\subsection{TREX is robust to most endoscopic image variations}

Performance of the baseline models and TREX models across longitudinal follow-up timepoints, where clinicians labeled the image at the last available follow-up (timepoint 0) and any other previous follow-ups are retrospectively assigned that label. TREX achieved the highest balanced accuracy and sensitivity at clinically relevant timepoints, particularly 3–6 months before clinical detection and at the final follow-up (0 months), outperforming single-image models (RN152-SI, SwinS-SI) and the conventional siamese baseline (CAT). Earlier timepoints (12–24 months before detection) remained challenging for all models, with performance near chance levels.

The impact of image artifacts (including blood, stool, telangiectasia, and poor image quality) was assessed by manually annotating images with these factors and then computing sensitivity and specificity (Fig.~\ref{fig:dataset_confounders}). Telangiectasia is a known clinical indicator of favorable treatment response~\cite{williams2024endoscopic, Safont2024, Van_der_Sande2021} and is therefore associated with a lower likelihood of tumor regrowth. In images containing telangiectasia, TREX showed a persistent imbalance between specificity ($\geq$ 80\%) and sensitivity ($\leq$ 50\%) across all timepoints (Fig.~\ref{fig:dataset_confounders}a). 

In the presence of blood, TREX maintained a more balanced specificity and sensitivity across all timepoints except at 6--12 months, indicating no systemic bias to detect LR in images containing blood (Fig.~\ref{fig:dataset_confounders}b). The absence of such bias is important as blood is not a clinical indicator of regrowth. By contrast, both stool and poor image quality may occlude or obscure tumors, and can occur in images corresponding to both LR and CR. In these subgroups, TREX showed a large imbalance in the earlier timepoints with higher specificity that diminished towards later timepoints for both stool (Fig.~\ref{fig:dataset_confounders}c) and poor quality (Fig.~\ref{fig:dataset_confounders}d). This trend was consistent with the higher balanced accuracy observed at later timepoints. Corresponding results for CAT and the single-image models are provided in \textit{SI Appendix}, Fig.~\ref{SI-Fig:SubLabels_AllModels}. Sample sizes for each image artifact can be found in \textit{SI Appendix}, Table~\ref{SI-tab:ImageArtifactNumbers}.

\subsection{TREX attention maps capture meaningful correspondences in images} 

We evaluated whether TREX extracts meaningful correspondences between features from the two timepoints despite a lack of spatial registration by visualizing the correspondences of the attention maps. Specifically, we visualized the query-to-target patch attentions on paired restaging and follow-up images (Fig.~\ref{Attention}). Across all examples, the model assigned high attention to anatomically corresponding image regions despite substantial differences in viewpoints, lighting, and tissue appearances. Fig.~\ref{Attention} shows query patches containing tumor abutting mucosal folds, with corresponding peak response localized to the lesion or scar. A similar pattern was observed for ulcerated regions, including areas containing blood (row 4 in Fig.~\ref{Attention}). These qualitative results suggest that the dual cross-attention mechanism is able to identify clinically relevant correspondences across longitudinal image pairs even in the absence of explicit spatial registration.

\begin{figure*}[!t]
    \centering
    \includegraphics[width=0.98\linewidth]{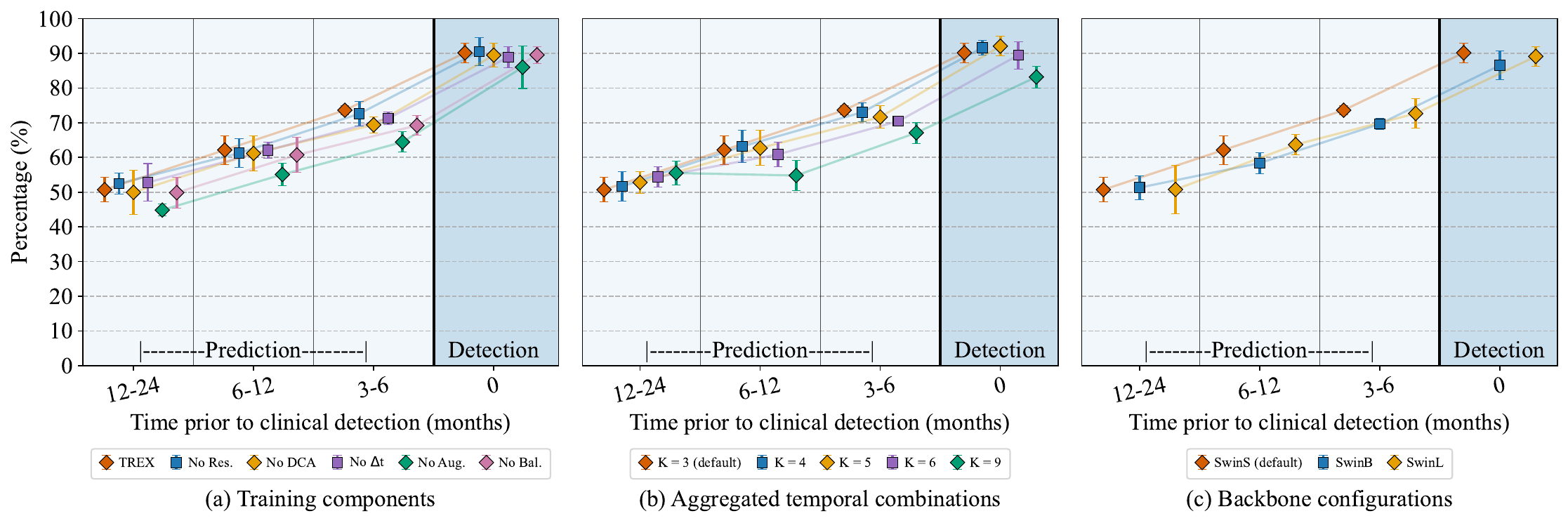}
    \caption{Ablation experiments evaluating the contribution of key TREX components and design choices. (a) Removing balanced sampling or data augmentation produced the largest reduction in balanced accuracy across all timepoints, while removing dual cross-attention (DCA) or temporal encoding ($\Delta t$) also consistently reduced performance, confirming the importance of both architectural and training components. (b) Performance of different top-$K$ aggregation strategies showed that $K = 3$ provided the most stable overall results, while larger $K$ values led to diminishing returns. (c) Comparison of Swin Transformer backbone sizes showed that the default Swin-Small backbone performed as well as or better than larger backbones across most timepoints.}
    \label{TREXComponents}
\end{figure*}

\subsection{Ablation experiments}

We performed ablation experiments to evaluate the impact of key model components, the choice of $K$ in top-$K$ prediction aggregation, and the size of the Swin Transformer backbone.

\subsubsection{Impact of different training components}

As shown in Fig.~\ref{TREXComponents}a, removing balanced sampling or data augmentation had the largest negative impact on performance, reducing balanced accuracy at all timepoints. Removing dual cross-attention (DCA) also reduced performance and had a slightly larger effect than removing temporal encoding ($\Delta t$) at earlier timepoints, although this difference diminished closer to clinical detection. Both of these components were important for accuracy, as seen by a decrease in accuracy across all analyzed timepoints compared to the TREX model. Together, these results indicate that both architectural and training components contribute meaningfully to TREX performance.

\subsubsection{Impact of K for top-K ensembles aggregation}

We evaluated the impact of varying the number of aggregated predictions $K$ in Fig.~\ref{TREXComponents}b. Increasing $K$ from 3 to 6 improved performance at the 12--24 months timepoint but resulted in comparable or degraded performance at the 6--12, 3--6, and 0 months (before clinical detection) timepoints. When $K$ was increased beyond 6, performance degraded substantially across most timepoints (except 12--24 months), indicating diminishing returns from averaging too many image pair combinations. Based on this observation, we used $K=3$ in all our experiments.

\subsubsection{Backbone size}

Fig.~\ref{TREXComponents}c shows the performance of different backbone sizes for TREX across longitudinal timepoints. The SwinS backbone performed as well as or better than the relatively larger backbones at most timepoints. Although SwinL showed a small improvement at the earliest timepoint, this advantage did not persist at later timepoints closer to clinical detection.

\section{Discussion}

We developed TREX, a longitudinal deep learning model that analyzes pairs of endoscopic images acquired at different timepoints and accurately detects rectal cancer regrowth. By combining cross-attention-based feature correspondence with temporal information encoding, TREX captures the spatial and temporal dynamics of the tumor site directly from unaligned images. Our approach is unlike prior works that rely on single timepoints~\cite{SPIE_Endo, williams2024endoscopic} or require spatial alignment~\cite{Chen2025IJROBP_PretreatmentMidtreatmentCT,Sun_LOMIAT_MICCAI2024,Ke2023DisColonRectum_DeepRPRC}, thereby making it well suited to real-world clinical workflows where consistent spatial registration is often not feasible. 

TREX demonstrated superior longitudinal performance compared to single-image models, maintaining 90\% balanced accuracy across follow-up intervals. Notably, it identified local regrowth 3 to 6 months ahead of clinical diagnosis (74\% balanced accuracy), highlighting a clear window for earlier intervention. Importantly, TREX achieved a clinical detection accuracy of  86.21\% comparable to that of expert clinicians (87.84\% $\pm$ 1.28\%). While its restaging AUC of 0.77 is similar to prior benchmarks of 0.76 to 0.83~\cite{Haak_Endo_ClinicVariables}, TREX uniquely allows for continuously evaluating and predicting treatment response, thereby allowing for progressive improvement in performance. Overall, these findings suggest that TREX is uniquely suited for both early response prediction and the detection of regrowth during active surveillance.

Analysis of common image confounders showed that TREX is largely invariant to non-diagnostic artifacts such as blood, stool, and poor quality in the images. However, TREX more often predicted complete response in images containing telangiectasia, a known marker of favorable treatment response~\cite{williams2024endoscopic, Safont2024, Van_der_Sande2021}. Furthermore, Grad-CAM analysis showed that TREX occasionally misclassified images containing nodules and normal mucosal folds. These instances likely reflects the inherent difficulty in longitudinal image pairing in presence of complex anatomical and non-anatomical variations, highlighting a specific target for future architectural refinement.

The current selection for organ preservation relies heavily on subjective assessment of endoscopic response, which is highly variable, reaching an accuracy between 65–85\%~\cite{CHINO20181247,EnodMRIForRestaging,felder2021} for detecting clinical response. Therefore, the need for an automated and objective assessment in patients undergoing watch-and-wait surveillance cannot be overstated~\cite{Williams2024}. While previous deep learning model have lacked the architectural complexity or sample size to match clinicians or detect local regrowth~\cite{thompson2023, Haak_Endo_ClinicVariables}, our study shows there is potential to achieve near-clinician-level assessment and  detect local regrowth. By striving to identify regrowth at its earliest stages, TREX offers an objective and accurate path to select patients for organ preservation and also ensures that those requiring curative surgery receive it without delay, thereby optimizing patient outcomes and enhancing their quality of life. 

Our work has a few limitations. We observed that performance varied with image pairings, with restaging-based pairs generally resulting in better results than arbitrary pairings. This suggests that the choice of reference timepoint may influence overall model performance. However, as our current approach only leverages pairs of timepoints rather than full longitudinal sequences, a systematic evaluation of pairing strategies (including filtering for image quality) remains to be thoroughly studied. Additionally, the current analysis predicts LR versus sustained CR at discrete follow-up timepoints rather than modeling time to regrowth directly. Given the highly variable response dynamics of rectal tumors~\cite{GarciaAguilar2011timing}, approaches that explicitly model the timing of regrowth may provide additional insights. Finally, prior clinician surveys have shown substantial inter-rater variability in the interpretation of endoscopic response features~\cite{felder2021}. Evaluating whether AI assistance improves consistency in this setting and exploring integration of other complementary modalities (such as MRI, histopathology, genomics, and clinical variables) remains an important direction for future work.


\section{Methods}
\subsection{Temporal Rectal Endoscopy Cross-Attention (TREX) approach overview}

\begin{figure*}[t]
\centering
\includegraphics[width=0.98\linewidth]{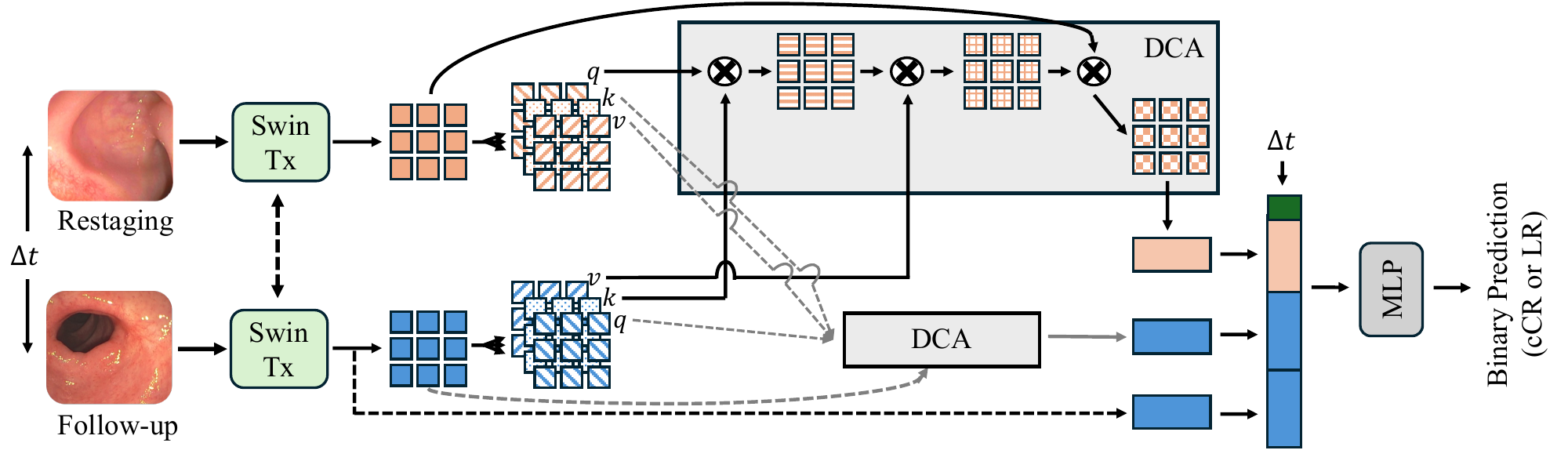}
\caption{Pairwise Temporal Rectal Endoscopy Cross-Attention architecture. Restaging (res) and follow-up (fup) images are processed through siamese Swin Transformer encoders. The final feature maps undergo dual cross-attention through two CA blocks to model temporal changes, followed by an MLP for classification of CR versus LR across variable follow-up timepoints ($\Delta t$).}
\label{fig:Architecture}
\end{figure*}

Fig.~\ref{fig:Architecture} depicts an overview of our pairwise temporal rectal endoscopy cross-attention architecture. The network backbone consists of a hierarchical shifted window (Swin) Transformer~\cite{liu2021swintransformerhierarchicalvision} used in a siamese network (SN) to extract features from two image timepoints, namely restaging $x_{res}$ and a follow-up scan $x_{fup}$. The network implements four encoder stages, and dual cross-attention (DCA) is applied to the Stage-4 features at $1/32$ spatial resolution, as shown in Fig.~\ref{fig:Architecture}. The features are represented as:
\begin{align}
F_{res} &= f^{(4)}_\theta(x_{res})\in\mathbb{R}^{\frac{H}{32}\times \frac{W}{32}\times C}
\label{eqn:f_pre}\\[6pt]
F_{fup} &= f^{(4)}_\theta(x_{fup})\in\mathbb{R}^{\frac{H}{32}\times \frac{W}{32}\times C}
\label{eqn:f_post}
\end{align}
\noindent where $H$ and $W$ denote the input image resolution and $C$ denotes the feature dimension at Stage 4. DCA uses two cross-attention modules to extract bidirectional information. Following cross-attention computation, the features are merged and processed through a multi-layer perceptron (MLP) to generate a binary classification of CR versus LR. Temporal information is incorporated by concatenating the normalized time difference $\Delta t$ with pooled feature representations prior to classification (at the classification head).

\subsubsection{Dual cross-attention (DCA)}

The endoscopic images acquired at the two timepoints are treated as independent sets of tokens, with no assumption of patch-level correspondences. DCA performs bidirectional attention by treating tokens from one image as queries and tokens from the other as keys and values. This symmetric computation allows the model to emphasize correlated features across the two images without requiring explicit spatial co-registration. The feature maps $F_{res}$ and $F_{fup}$ have spatial resolution $H/32 \times W/32 \times C$. Cross-attention is computed by flattening the spatial dimensions so that both $F_{res}$ and $F_{fup}$ are reshaped into matrices of size $(H/32 \cdot W/32)\times C$, and then applying multi-head attention with learned projections $W_q, W_k, W_v \in \mathbb{R}^{C \times C}$ in each direction as:

\begin{align}
q_{res} &= F_{res}W_q, \hspace{0.8em} k_{fup} = F_{fup}W_k, \hspace{0.8em} v_{fup} = F_{fup}W_v \\
A_{res} &= \sigma\left(\frac{q_{res} k_{fup}^T}{\sqrt{d_h}}\right), \quad \text{CA}(F_{res}) = A_{res} v_{fup}
\end{align}
\noindent, where $d_h = C/h$ for $h=8$ attention heads. The symmetric operation computes $\text{CA}(F_{fup})$ using $F_{res}$ as keys and values:
\begin{align}
q_{fup} &= F_{fup} W_q, \quad k_{res} = F_{res} W_k, \quad v_{res} = F_{res} W_v \\
A_{fup} &= \sigma\left(\frac{q_{fup} k_{res}^T}{\sqrt{d_h}}\right), \quad \text{CA}(F_{fup}) = A_{fup} v_{res}
\end{align}

Layer normalization (LN) is then applied to the features to stabilize training:
\begin{equation}
H_{res} = \text{LayerNorm}(F_{res} \odot \text{CA}(F_{res}))
\end{equation}
\begin{equation}
H_{fup} = \text{LayerNorm}(F_{fup} \odot \text{CA}(F_{fup}))
\end{equation}

The attention-guided features $H_{res}$ and $H_{fup}$ are combined using a residual connection with the Stage-4 features $F_{fup}$ to stabilize training.

\subsubsection{Temporal encoding and classification}

Next, to further emphasize the differences between early versus later follow-ups, a time-conditioned representation was extracted by incorporating temporal information explicitly to the model. This was accomplished by computing the time difference $\Delta t$ (in days, normalized to $[0,1]$ over a 2-year span) between the restaging and follow-up images using metadata, and concatenated with the features following global average pooled (GP) features applied to $H_{res}$ and $H_{fup}$, and $F_{fup}$ and then combined with the multi-layer perceptron (MLP) used for classification as:
\begin{equation}
z = \sigma(\text{MLP}([(\text{GP}(H_{res}), \text{GP}(H_{fup}), \text{GP}(F_{fup}), \Delta t])
\end{equation}
where $z \in \mathbb{R}^2$ represents class probabilities for CR and LR and $\sigma$ is the sigmoid activation producing a probability of tumor regrowth. The temporal encoding enables the model to account for the expected rate of tumor progression and the reliability of visual observations based on image quality. 

\subsubsection{Rank aggregated prediction}
Multiple images are typically acquired at each study timepoint. Hence, all available images were utilized by combining restaging images with images from a given follow-up to generate an ensemble of predictions. A top-$K$ strategy, with $K = 3$ (the most common number of images acquired at a study timepoint), was used to select predictions with the highest probabilities. We found this approach to be more resilient to imaging variations than selecting the single best candidate image pair with the highest probability. 

\subsection{Datasets}

All datasets were split into five folds at the patient level to avoid data leakage, with four folds used for training and one fold used for validation in each iteration. The details of the total number of images and pairs for each experiment and split are available in \textit{SI Appendix}, Table~\ref{SI-tab:fold_splits}.

\indent \textit{a) Primary analysis} used 3,326 endoscopic images from 197 patients acquired using white-light flexible endoscopy with an Olympus scope (model CF-160S). The images varied in size from 568 $\times$ 424 pixels to 1920 $\times$ 1080 pixels and were rescaled to 224 $\times$ 224 pixels for all models. Images were acquired at restaging appointments performed within 8 weeks of completing TNT and during follow-up examinations every 3--6 months in the WW period until clinical detection (referred to as `0 months before last follow-up'). Images at intermediate follow-ups were retrospectively assigned the LR label, even when no tumor was visible (n = 1,389), if LR was diagnosed at clinical detection. Conversely, images at intermediate follow-ups were assigned the CR label if a sustained CR was confirmed at clinical detection (n = 1,937). More images corresponded to CR because these patients typically had longer follow-up than patients with LR.

\indent \textit{b) Secondary analysis} was performed to evaluate the model's capability to predict LR versus CR from pre-TNT images alone, restaging images alone, and the combination of pre-TNT and restaging images. This analysis was designed to assess the benefit of using temporal information to predict LR at the earliest possible timepoint relative to single timepoint images. This dataset consisted of 488 endoscopic images from 101 patients.

\subsection{Implementation details}

The Swin Transformer variant was a Swin-Small with a patch size of 4 $\times$ 4 and a window size of 7 $\times$ 7 for inputs of 224 $\times$ 224 pixels. Pretrained ImageNet weights~\cite{Deng2009}, as provided by PyTorch~\cite{paszke2019pytorch}, were used to initialize the backbone before fine-tuning. The classification head consisted of a linear layer, ReLU activation, and dropout, followed by a final linear layer that outputs logits, which are converted into tumor-regrowth class probability via softmax. 

All models were fine-tuned with the Adam optimizer~\cite{kingma2015adam} with a learning rate of 2$\times$ 10$^{-4}$, combined with a linear rate scheduler with 10 epochs of warmup. Training used cross-entropy loss with a batch size of 8 for 30 epochs. Data augmentation included random image rotations in 90$^\circ$ steps, random horizontal flips, and random vertical flips, all of which were used to address data limitations for colonoscopy image analysis~\cite{ramesh2023dissecting}. Balanced sampling was used to reduce the class imbalance between the CR and LR within the mini-batches at each iteration. 

To obtain a reliable ensemble within the low-data regime, we performed stratified 5-fold cross-validation. For each fold, the model performance was recorded at last epoch, and the results were aggregated. The reported metrics correspond to the mean and standard deviation across the five folds. All experiments used PyTorch (v1.13.1) with NVIDIA GPUs.

\subsection{Model baselines}

TREX was compared against three baseline models: (a) a Swin-Small single-image model (SwinS-SI), which used the same transformer encoder as TREX but processed single-images; (b) a convolutional-network-based ResNet-152 single-image model (RN152-SI); and (c) a Swin-Small siamese network that concatenated Stage-4 features, denoted CAT. All models were trained and tested with identical datasets with inference performed using top-$K$ strategy as described previously.

\subsection{Experiments}
Experiments were performed to measure: (a) the accuracy of distinguishing LR from CR at clinical detection versus retrospective early prediction at 3--6, 6--12, and 12--24 months before clinical detection, (b) robustness to imaging variations, (c) clinical validation through comparison with clinician assessments, (d) assessment of feature-map separability through unsupervised clustering and attention-map analysis, (e) the impact of model design choices through ablation experiments, and (f) the advantage of temporal image combinations at the earliest post-treatment timepoint (restaging), using a subset of scans with pre-TNT and restaging images compared against single-image models. 

\subsection{Metrics}
Accuracy was reported using balanced accuracy, sensitivity, and specificity, with a threshold of 0.5 for the binary classification. Ground truth labels for LR versus CR were available only at the last follow-up, defined as the visit where a physician decided either to continue watch-and-wait (no visible LR), confirmed a sustained CR and concluded the watch-and-wait (more than 2 years of follow-up), detected an LR confirmed following surgery, or recorded if the patient was lost to follow-up. This last visit served as 0 months reference point, and performance was summarized in four non-overlapping time bins relative to last visit: 12--24 months, 6--12 months, 3--6 months, and 0 months. Intermediate follow-up scans did not have ground truth labels beyond visual clinical assessment. Therefore, we used the \textit{gold-standard} ground truth obtained at the last follow-up. This means that applying the model to intermediate follow-ups constitutes a prediction task, whereas applying it to the last follow-up constitutes a detection task.

\subsection{Clinical validation with surgeon survey}

TREX, CAT, and SwinS-SI were evaluated against 16 experts, including colorectal surgeons and surgical trainees, using an anonymized REDCap survey comprising 58 de-identified endoscopic images from the last available follow-up, when ground truth labels were available (CR: 31, LR: 27)~\cite{Williams2024}. In addition to reporting model performance on these 58 images, we also assessed TREX and CAT using all corresponding prior image pairs available for the same patients. All participants had experience with WW either through dedicated research years or clinical expertise. Surgeons were blinded to patient clinical outcomes and were asked to determine the presence or absence of tumor in each individual image. Each respondent's performance was evaluated on the survey images, and group averages were calculated by experience level, including surgical residents, colorectal fellows, and attending colorectal surgeons.

\subsection{Statistical Analysis}

The impact of using TREX was assessed at the patient level by evaluating its accuracy against RN152-SI, SwinS-SI, and CAT. To prevent data leakage, cross-validation folds were split strictly at the patient level, ensuring that no patient contributed images to both the training and test sets of any fold. For each patient, the top-$K$ predictions from the cross-validated models were aggregated and combined via majority voting to produce a single patient-level prediction.

Statistical significance of the differences in patient-level accuracy between TREX and each baseline was assessed using McNemar's test~\cite{mcnemar1947}, applied to the paired patient-level predictions. A p-value below 0.05 was considered statistically significant. The odds ratio quantifies the magnitude of this difference: it is computed as the ratio of discordant pairs (patients correctly classified by TREX but misclassified by the baseline, divided by patients misclassified by TREX but correctly classified by the baseline). An odds ratio greater than 1 indicates that TREX is more likely than the baseline to correctly classify patients on which the two models disagree, with larger values reflecting a stronger advantage.


\section{Compliance with Ethical Standards}
This retrospective research study was conducted in line with the principles of the Declaration of Helsinki. Approval was granted by the Ethics Committee of Memorial Sloan Kettering Cancer Center.

\section{Acknowledgments}

This research was partially supported by the Department of Surgery at Memorial Sloan Kettering. We thank Maria Widmar, Iris H. Wei, Emmanouil P. Pappou, Garrett M. Nash, Martin R. Weiser, and Philip B. Paty, along with Hannah Thompson, Hannah Williams, J. Joshua Smith, and Julio Garcia-Aguilar, for their assistance in collecting endoscopic images. Additionally, Francisco Sanchez-Vega acknowledges support through a research grant from The Society of MSK.

\section{Author contributions}
J.T.G., A.R., M.R.S., J.G., and H.V. designed research; J.T.G., D.K., A.R., H.W., H.T., C.L., and H.V. performed research; J.T.G, D.K., H.W., H.T., C.L., J.J.S., and J.G. collected data; J.T.G., D.K., A.R., H.W., H.T., C.L., J.G. and H.V., analyzed data; and J.T.G., D.K., A.R., J.J.S., F.S.V., M.R.S., J.G. and H.V. wrote the paper; M.R.S. advised on experiments; J.G. secured funding.

\section{Competing interests}
The authors declare no competing interests.

\bibliography{references}

\clearpage


\end{document}